\documentclass[letterpaper]{article} 
\usepackage{aaai23}  
\usepackage{times}  
\usepackage{helvet}  
\usepackage{courier}  
\usepackage[hyphens]{url}  
\usepackage{graphicx} 
\usepackage{natbib}  
\usepackage{caption} 
\usepackage{algorithm}
\usepackage{algorithmic}

\usepackage{newfloat}
\usepackage{listings}

\DeclareCaptionStyle{ruled}{labelfont=normalfont,labelsep=colon,strut=off} 
\floatstyle{ruled}
\newfloat{listing}{tb}{lst}{}
\floatname{listing}{Listing}
\usepackage{amsmath,amsfonts,bm}

\def\eqref#1{equation~\ref{#1}}

\def\1{\bm{1}}

\DeclareMathAlphabet{\mathsfit}{\encodingdefault}{\sfdefault}{m}{sl}
\SetMathAlphabet{\mathsfit}{bold}{\encodingdefault}{\sfdefault}{bx}{n}

\def\gC{{\mathcal{C}}}

\def\gE{{\mathcal{E}}}

\def\gG{{\mathcal{G}}}

\def\gL{{\mathcal{L}}}

\def\gV{{\mathcal{V}}}

\newcommand{\E}{\mathcal{E}}

\newcommand{\R}{\mathbb{R}}

\usepackage{subfigure}
\usepackage{amsmath}
\usepackage{amssymb}

\usepackage{enumitem}

\newcommand{\aggregate}{\mathrm{agg}}
\renewcommand{\E}{\mathcal{E}}
\newcommand{\GNN}{\mathsf{GNN}}
\newcommand{\M}{\mathcal{M}}

\renewcommand{\R}{\mathbb{R}}

\newcommand{\trace}{\mathrm{trace}}

\newtheorem{definition}{Definition}[section]
\newtheorem{theorem}{Theorem}[section]

\newtheorem{assumption}{Assumption}[section]

\newtheorem{corollary}{Corollary}[section]

\title{USER: Unsupervised Structural Entropy-based Robust Graph Neural Network}

\author {
    Yifei Wang\textsuperscript{\rm 1},
    Yupan Wang\textsuperscript{\rm 1}, 
    Zeyu Zhang\textsuperscript{\rm 1}, 
    Song Yang\textsuperscript{\rm 1}, 
    Kaiqi Zhao\textsuperscript{\rm 1}, 
    Jiamou Liu \textsuperscript{\rm 1}\thanks{Corresponding author}
}
\affiliations {
    \textsuperscript{\rm 1} School of Computer Science, The University of Auckland, New Zealand\\
    \{wany107, ywan980, zzha669,
    syan382\}@aucklanduni.ac.nz,
    \{kaiqi.zhao,
    jiamou.liu\}@auckland.ac.nz
}

\usepackage{bibentry}

\begin{document}

\maketitle

\begin{abstract}
Unsupervised/self-supervised graph neural networks (GNN) are vulnerable to inherent randomness in the input graph data which greatly affects the performance of the model in downstream tasks. 
In this paper, we alleviate the interference of graph randomness and learn appropriate representations of nodes without label information. To this end, we propose USER, an unsupervised robust version of graph neural networks that is based on structural entropy. We analyze the property of intrinsic connectivity and define intrinsic connectivity graph. We also 
identify the rank of the adjacency matrix as a crucial factor in revealing a graph that provides the same embeddings as the intrinsic connectivity graph. We then introduce structural entropy in the objective function to capture such a graph. 
Extensive experiments conducted on clustering and link prediction tasks under random-noises and meta-attack over three datasets show USER outperforms benchmarks and is robust to heavier randomness.
\footnote{Full proof, experimental details, and code of our work is available at \url{https://github.com/wangyifeibeijing/USER}.}
\end{abstract}

\section{Introduction}
\begin{figure*}[h]
		\centering
		\includegraphics[width=17cm]{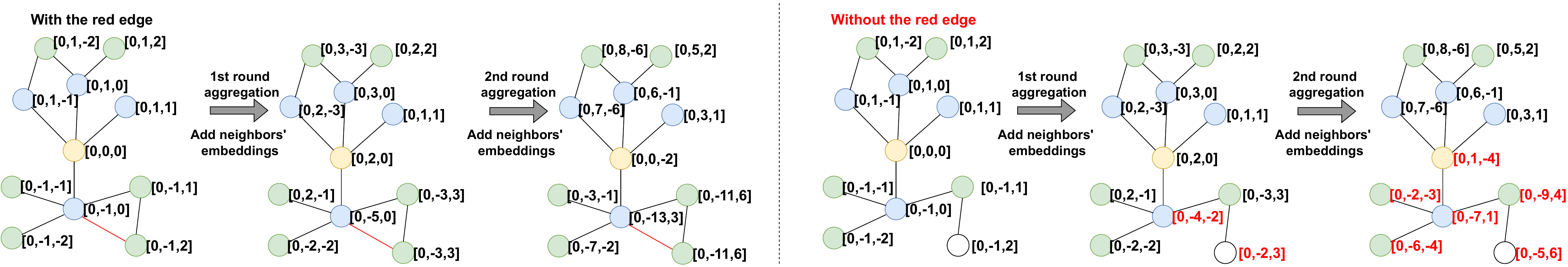}
		\caption{Modifying an edge causes more than half of the nodes' embeddings to change after two rounds of aggregation}
	    \label{fig:Modifying an edge}
	\end{figure*}
Neural-based methods for processing complex graph data have become indispensable to a wide range of application areas from social media mining, recommender systems, to biological data analysis and traffic prediction. 
{\em Graph representation learning} (GRL) plays a central role in these methods, providing vectorized graph encodings which are crucial for downstream tasks such as community detection, link prediction, node classification, and network visualization \cite{hamilton2017representation}. 
Among the many GRL methods that emerged in recent years, {\em graph neural network} (GNN) \cite{hamilton2017inductive,kipf2017semi,velickovic2018graph} provides a powerful paradigm that extracts graph encodings through a recursive aggregation scheme \cite{kipf2017semi}. 
The aggregation scheme learns a node's embedding using both the feature of the node itself and aggregated feature of its neighbours, thereby capturing structural information of the graph.
The advantage of GNN-based models has been attested by outstanding performance on many tasks \cite{kipf2016variational,wang2017mgae,PanHLJYZ18,GaoH18b,velickovic2019deep,zhang2019attributed,wang2019attributed,pan2019learning,cui2020adaptive,MavromatisK21}. 

Despite the successes above, the performance of a GNN-based model hinges on reliable input graph data \cite{jin2020graph,chen2020survey,luo2021learning}.
More specifically, small perturbation on the input graph may ends in drastically different encodings after the recursive aggregation scheme (see Figure~\ref{fig:Modifying an edge}).
However, The randomness of input graph is inevitable.
As \cite{wang2019learning} discussed, the edges in the graph are formed randomly, following an underlying intrinsic connectivity distribution of nodes.
\cite{fortunato2010community} asserts that such distribution is highly inhomogeneous, with edges concentrated within some conglomerates of nodes, resulting in a community structure. Each such conglomerate is called a community: they have dense inner-connections but sparse inter-connections.   
The community structure of a network can be interpreted in a probabilistic way: A fixed but unknown edge (probabilisty) distribution is presented between any pairs of nodes that determines a community structure, yet we only observe a {\em sample} from this distribution.  
%
This inherent randomness triggers the perturbations in the input graph that interferes the recursive aggregation scheme. 
%


It is therefore desirable to develop a model that captures the so-called ``{\em intrinsic connectivity graph}'', a graph that reflects the intrinsic connectivity in the dataset.
A straightforward approach is to unravel the intrinsic connectivity graph from node labels that indicates the communities.
However, ground truth labels are often not available in real-world applications. Our task is thus to develop an {\em unsupervised} approach for unsupervised/self-supervised GNN. 
For this task, one needs to address the following three challenges:
\textbf{(a)}
The first challenge demands an operational criterion for alleviating the interference of graph randomness\footnote{Here by graph randomness, we only consider the perturbation of edges.}. 
Since 
ground truth labels are not present, we need to find a new criterion to find a graph that mitigates the interference of graph randomness.
 \textbf{(b)} Given such a criterion in (a), the second challenge concerns how a graph that meets it can be learnt. More precisely, this challenge seeks for an objective function that guides to reveal a new graph that satisfies the criterion. 
 \textbf{(c)} The third challenge seeks a framework to generate the new graph.
Common GNN models generate graphs with the node embedding results.
However, \cite{jin2020graph} assert that graphs generated from interfered node embeddings are unreliable.
A desirable solution to this question would be learning embeddings and the new graph simultaneously.

For {\bf (a)}, 
we will show in {\bf Section~\ref{sec:Criteria for Mitigating Randomness}} that 
there exist multiple ``innocuous graphs'' for which GNN may produce the same embeddings as the desired intrinsic connectivity graph. We call these graphs {\em GNN-equivalent}.   
Thus any one of such innocuous graphs can help GNN mitigate the interference of randomness. 
As the number of groups in the intrinsic connectivity graph (the number of communities $c$) is known for many datasets,
we justify two assertions about the innocuous graphs: 
For the innocuous graphs, the rank of its corresponding adjacency matrix is no less than the number of groups in the intrinsic connectivity graph. 
Then, if we partition innocuous graphs into groups with high concentrations of edges inside groups, and low concentrations between them,
features of two nodes that in the same group should be relatively similar. 
These assertions direct our pursuit for innocuous graph. 

For {\bf (b)}, to reflects all assertions above, we develop a tool to learn a graph that satisfies the conditions above. In {\bf Section~\ref{sec:The objective function based on structural-entropy}}, we invoke  structural information theory \cite{li2016structural,Li2016ResistanceAS,liu2019rem}. Through a class of {\em structural entropy} measures, structural information theory recently emerges to capture intrinsic information contained within a graph structure and have been increasingly applied to graph learning.  
Here, we connect the notion of network partition structural information (NPSI), a type of structural entropy, with the rank of the adjacency matrix and show that minimizing structural entropy would facilitate the search for a innocuous graph.  

For {\bf (c)}, we combine the tools developed above and design a novel framework, called {\bf U}nsupervised {\bf S}tructural {\bf E}ntropy-based {\bf R}obust graph neural network (USER),  to support GNN with a trainable matrix to learn the adjacency matrix of an innocuous graph. See {\bf Section~\ref{sec:USER-structure}}. Our method makes it possible to learn the innocuous graph structure and the node embeddings simultaneously. As the embeddings are derived from the innocuous graph, rather than the input graph, they are tolerant to randomness. 

We developed a series of experiments to validate our USER framework. 
Thee experiment results show that on both clustering and link prediction tasks, with the support of USER, even traditional GAE model surpasses state of the art baselines. We inject randomness into the graph.
With $50\%$ random noises, accuracy of USER improves from SOTA by up to $14.14\%$ for clustering on well-known Cora dataset, while improvement for link prediction reaches $13.12\%$ on  Wiki dataset. 
Moreover, USER exhibits higher advantage in the presence of adversarial attacks.
Facing $20\%$ meta-attack \cite{zugner2019adversarial}, USER's improvements over SOTA is up to $190.01\%$  for clustering on Citeseer. Our contributions can be summarized as follows:
\begin{itemize}
    \item We are the first to introduce the notion of GNN-equivalent and innocuous graph. We utilize them for mitigating the interference of graph randomness;
    \item We proposed a structural entropy-based objective function that is suitable for learning the innocuous graph;
    \item We conduct extensive experiments, which show that USER performs effectively confronting randomness.
\end{itemize}


\section{Related Work}
\label{sec:Related Work}

\noindent \textbf{GRL and GNN.}
Graph representation learning (GRL) generates vectorized encoding from graph data. 
Nowadays GRL is instrumental in many tasks that involves the analysis of graphs \cite{hamilton2017representation}. 
As a mainstream GRL paradigm, graph neural network (GNN)  captures a node's structural information and node features by recursively aggregating its neighborhood information using an {\em information aggregation scheme}. Based on this idea, Graph Autoendoer (GAE) and Variational GAE (VGAE) \cite{kipf2016variational} are developed to use GCN \cite{kipf2017semi} as an encoder to learn node embeddings, and an inner product decoder to reconstruct the graph structure.
As a variant of GAE, ARGA \cite{PanHLJYZ18} trains an adversarial network to learn more robust node embeddings. To alleviate the high-frequency noises in the node features, Adaptive Graph Encoder (AGE) \cite{cui2020adaptive} utilizes a Laplacian smoothing filter to prepossess the node features.
Different from reconstructing the graph structure, 
maximizing the \textit{mutual information} (MI) \cite{hjelm2018learning} is another well-studied approach for GRL. 
For example, 
the model DGI \cite{velickovic2019deep} employs GNN to learn node embeddings and a graph-level embedding, then maximize MI between them to improve the representations’ quality.
The model GIC \cite{mavromatis2020graph} follows this idea and seeks to additionally capture community-level information of the graph structure.

Despite GNN's outstanding performance, studies have demonstrated that small perturbation on the input graph can fool the GNN \cite{li2018adaptive,zhu2021deep,wan2021graph,wu2019adversarial}. 
These perturbations are inevitable, especially for unsupervised models.
New learning paradigms are proposed to alleviate the influence of such perturbations.
The model Cross-graph \cite{wang2020cross} maintains two autoencoders. Every encoder learns node embeddings and reconstructs adjacency matrix to be passed to the peer-autoencoder as the input for the next iteration. 
{\em Graph contrastive learning (G-CL)} models such as GCA \cite{zhu2021graph} also improve the robustness of GNN.
These methods construct data augmentation and negative pairs by modifying the input graph structure.
However, none of these works explain how these perturbations were formed.
In this paper, inspired by \cite{wang2019learning,fortunato2010community,zhang2019attributed,wu2019simplifying,zhu2020simple} we introduce the notion of innocuous graph to learn the embeddings same to these corresponding to the intrinsic connectivity graph, which helps GNN models to mitigate the impact of randomness.

\noindent \textbf{Structural entropy.}
Structural entropy is a major tool utilized in our paper.
An entropy measure has long been sought after in computer science to analyze the intrinsic information embodied in structures \cite{brooks2003three}. 
Several classical entropy measures have been designed for this purpose \cite{dehmer2008information,anand2009entropy}. 
In particular, the model infomap \cite{rosvall2009Infomap} tries to analyze graphs with a form of entropy defined on  random walks. 
In \cite{li2016structural, Li2016ResistanceAS}, the authors re-invented structural information theory and proposed a hierarchy of {\em  structural entropy} to analyze networks. This notion has been utilized in several work, e.g., \cite{liu2019rem,liu2022residual,chen2019distributed}, to adversarial graph learning. 
However, to date, no study has attempted to integrate structural entropy with enhance GNN's resilience to randomness in graph data. 

\begin{figure*}[h]
	\centering
	\includegraphics[width=17cm]{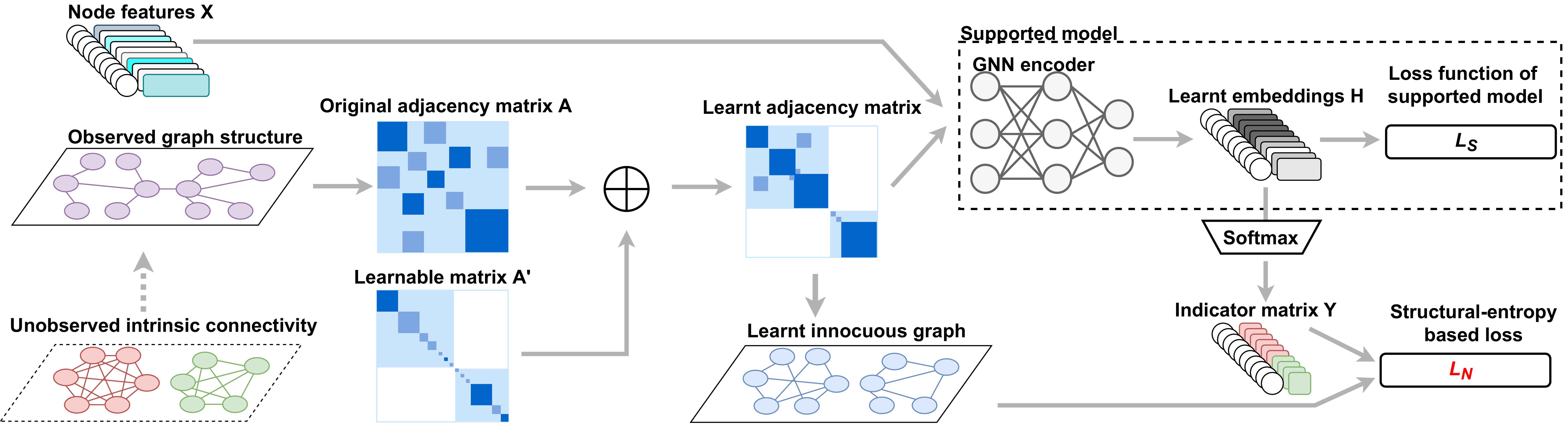}
	\caption{The USER framework. To mitigate randomness-interference in observed graph, an innocuous graph is constructed. We optimize structural entropy-based $\gL_N$ to the learn innocuous graph.}
	\label{The Structure of USER}
\end{figure*}

\section{Criteria to Mitigate Randomness}\label{sec:Criteria for Mitigating Randomness}
We use $\vec{x},\vec{y},\ldots$ to denote vectors where $x_i$ denotes the $i$th entry of $\vec{x}$. 
We use capital letters $X,Y,A,\ldots$ to denote real-valued matrices. For any matrix $M$, $M_i$ denotes the $i$th row vector and $M_{ij}$ denotes the $(i,j)$th entry of $M$.
In this paper, we focus on undirected, unweighted graphs where every node is associated a $d$-dimensional feature vector. Formally, such a graph can be denoted by $\gG=\left(\gV, \E, X\right)$ where $\gV$ is a set of $n$ nodes $\{v_1,\ldots,v_n\}$, $\E$ is a set of edges $\{v_i,v_j\}$, and $X\in \M_{n,d}(\R)$ denotes the {\em feature matrix} where $X_i$ is the feature vector of node $v_i$. 
The pair $(\gV, \E)$ is represented by an {\em adjacent matrix} $A\in \M_n(\{0,1\})$, where $A_{ij}=1$ if $\{v_i,v_j\}\in \E$.
Here we assume the graph does not contain any isolated node. Indeed, most studies on GNN omit isolated nodes before training \cite{kipf2016variational,MavromatisK21}.
At last, we use $\gC_0,\gC_1,\ldots \gC_{c-1}$ to denote $c$ sets of nodes. 
If they satisfy:  $k\neq m \Rightarrow \gC_k\cap \gC_m =\varnothing$ and $\forall k <c\colon \gC_k\neq \varnothing$ , we call them partitions.

Taking input graph $\gG$, a {\em graph neural network (GNN)} can be denoted by the function 
\begin{equation}       
\label{eq:GNN}
\begin{small}
		\GNN\left(A,X,\left\{W^{(\ell)}\right\}\right)=H^{(t)}\\
\end{small}
\end{equation} 
where $H^{(0)}=X$ and 
\[
H^{(\ell)}=\sigma(\aggregate(AH^{(\ell-1)}W^{(\ell)})) \text{ for all $\ell \in (0,t]$},
\]
$H^{(\ell)}\in\R^{n\times d^\ell}$, $W^{(\ell)}\in\R^{d^{(\ell-1)}\times d^{(\ell)}}$, and $d^{(0)}=d$. 
%
Here $H^{(\ell)}$ is the matrix learned by the $\ell$th information aggregation layer and $H^{(0)} = X$, taking the original features as the input to the $1$st layer;
$\sigma(\cdot)$ is the activation function; $\aggregate(\cdot)$ is the aggregation; and $W^{(\ell)}$ contains learnable parameters. 
GNN with non-injective $\sigma(\cdot)$ and $\aggregate(\cdot)$ are inefficient when learning graph structures  \cite{xu2018how}. Thus we only discuss GNN with injective $\sigma(\cdot)$ and $\aggregate(\cdot)$ functions. 
By (\ref{eq:GNN}), in GNN models, the vector representation of a node is computed  with not only its own features but also features of its neighbors accumulated recursively.

As mentioned above, 
a real-world input graph dataset is inherently random and unstable \cite{jin2020graph}. On the other hand, such datasets would reflect certain hidden but stable underlying {\em intrinsic connectivity} distribution \cite{wang2019learning, fortunato2010community}.
\cite{fortunato2010community} asserts that for a dataset which can be naturally separated, say, into $c$ partitions (or {\em classes} in a node classsificial task), intrinsic connectivity satisfies that nodes in the same partition are more likely to be connected than nodes in different partitions. We capture this intrinsic connectivity with the next definition. 
\begin{definition}[Intrinsic connectivity graph]
	\label{def:Intrinsic connectivity graph}
	For a dataset that contains $c$ partitions,
	suppose $\gG_{I}=(\gV, \E_{I})$ satisfies:
	For any two nodes $v_i$ and $v_j$, there exists an edge $(v_i,v_j)\in \E_I$ iff $v_i$ and $v_j$ belong to the same partition.
	We call $\gG_{I}$ the {\em intrinsic connectivity graph}. 
\end{definition}
Let $Rank(M)$ denote the rank of a matrix $M$. 
\begin{theorem}[Rank of $\gG_{I}$'s adjacency matrix $A_I$]
	\label{thm:Rank of GI's adjacency matrix AI}
	For a dataset that contains $c$ partitions, we have $Rank (A_I) = c$ where $A_I$ is $\gG_{I}$'s adjacency matrix.
\end{theorem}
Our aim is to extract a new graph from a real-world dataset to mitigate the interference of graph randomness. Without ground truth label, finding this intrinsic connectivity graph is impractical.
However, we observe that, GNN may learn the same embeddings from different input graphs:
\begin{definition}[GNN-equivalent]
	\label{def:GNN-equivalent} Let $\gG_0=(\gV, \E_0, X)$ and $\gG_1=(\gV, \E_1, X)$ be two graphs with the same set of nodes and adjacency matrices $A_0$ and $A_1$, respectively.
	Suppose we run GNN respectively on these two graph, and the following holds:  for any feature matrix  $X$.
	in each layer $\ell$, and any $W_0^{(\ell)}$, there exist weights $W_1^{(\ell)}$ such that:
	\[
        \sigma(\aggregate(A_0H^{(\ell-1)}W_0^{(\ell)}))  = \sigma(\aggregate(A_1H^{\ell-1)}W_1^{(\ell)})).
    \]
	Then we call $\gG_0$ and $\gG_1$ {\em GNN-equivalent}.
\end{definition}
By Def.~\ref{def:GNN-equivalent}, when $\gG_1$ is GNN-equivalent to $\gG_0$, a GNN with $\gG_1$ as input may learn the same embeddings as if  $\gG_0$ is the input.
Thus using graphs GNN-equivalent to intrinsic connectivity graph $\gG_I$ makes it possible for the GNN to learn the same embeddings as inputting $\gG_I$. We call such a graph {\em innocuous}. 
\begin{definition}[innocuous graph]
	\label{def:Innocuous graph}
	Suppose $\gG_{I}$ is the intrinsic connectivity graph for a dataset. 
	An {\em innocuous graph} $\gG'$ is one that is GNN-equivalent to $\gG_{I}$.
\end{definition}
To search for such graphs, we introduce the necessary condition for being GNN-equivalent to a specific graph:
\begin{theorem}[necessary condition of GNN-equivalence]
	\label{thm:Necessary condition for GNN-equivalent}
	$\gG_1$ is GNN-equivalent to $\gG_0$ only if  \begin{small}
	   $Rank(A_1) \geq Rank(A_0)$.
	\end{small}
\end{theorem}

\begin{corollary}[necessary condition of innocuous graph]
	\label{cor:Necessary condition of being innocuous graph}
	$\gG'$ is a innocuous graph only if  $Rank(A') \geq Rank(A_I)$.
\end{corollary}
By Theorem~\ref{thm:Rank of GI's adjacency matrix AI} and Corollary~\ref{cor:Necessary condition of being innocuous graph}, adjacency matrix $A'$ of innocuous graph $\gG'$ satisfies $Rank(A') \geq c$. 

Aside from the property above, we further remark on another commonly-used assumption \cite{wu2019adversarial,jin2020graph}: {\em In a graph over which a GNN may extract  semantically-useful node embeddings, adjacent nodes are likely to share similar features than non-adjacent nodes.} This formulation, however, only considers information aggregation of GNN along a single edge. We now extend feature smoothness to group-level. Let $f(\vec{x},\vec{y})$ be a function that evaluates similarity between learnt node embeddings, i.e., similarity between two embedding vectors $\vec{x}$ and $\vec{y}$ leads to a smaller $f(\vec{x},\vec{y})$.  We formulate group-level feature smoothness of a innocuous graph:
\begin{assumption}[group-level feature smoothness]
	\label{assump:Group-level feature smoothness}
	Suppose $k\neq m$. Then  for any three nodes $v_a,v_b,v_c$ that satisfy $v_a\in\gC_k$, $v_b\in\gC_k$ and $v_c\in\gC_m$, we have $f(X_a,X_b)\leq f(X_a,X_c)$.
\end{assumption}
In the next section, we formulate an overall criterion for finding an innocuous graph, which incorporate a necessary condition (Corollary~\ref{cor:Necessary condition of being innocuous graph}) and an auxiliary assumptions (Assumptions~\ref{assump:Group-level feature smoothness}).

\section{Structural Entropy-based Loss}
\label{sec:The objective function based on structural-entropy}
As discussed above, our model need to learn a graph that satisfies necessary conditions (Corollary~\ref{cor:Necessary condition of being innocuous graph} and Assumptions~\ref{assump:Group-level feature smoothness}) for obtaining an innocuous graph.
In this section, we interpret these conditions using the language of structural information theory and formulate an optimization problem.
Following recent progress on structural information theory \cite{li2016structural}, we invoke the notion of {\em network partition structural information (NPSI)}, which was not be used in GNN models before.

To explain NPSI, we firstly introduce the following notations: $P(\gG)=\{\gC_0,\gC_1,\ldots \gC_{r-1}\}$ is a {\em partition} of $\gG$. 
Then, $P(\gG)$ can be denoted by a matrix $Y\in \{0,1\}^{n\times r}$, where $Y_{ik} =1$ if $v_i \in \gC_k$ otherwise $Y_{ik} = 0$. We call $Y$ the {\em indicator matrix}. Since $\gC_k \neq \varnothing$, $(Y^TY)_{kk} > 0$, and since $\forall k\neq m$, $\gC_k \cap \gV_m = \varnothing$, if $k\neq m$, $(Y^TY)_{km} = 0$.

For a graph $\gG$ and partition $P(\gG)$, let $vol_k$ be the number of edges with at least one node in $\gC_k$ and $g_k$ be the number of edges with only one node in $\gC_k$. Then by \cite{liu2019rem}, NPSI is:
\begin{equation}
\label{eq:NPSI node form}
\begin{small}
    \begin{aligned}
        &NPSI_{\gG P(\gG)}=\sum_{k<r}\left(\frac{vol_k-g_k}{2|\gE|}\log_2\frac {vol_k}{2|\gE|}\right)
    \end{aligned}
\end{small}
\end{equation}
To utilize it in GNN models, we define a matrix form of NPSI.
Note that $vol_k-g_k$ is the number of edges  with both nodes in $\gC_k$, which equals to the $k$-th diagonal element in $Y^TAY$, 
while the $k$-th value in sum of column in $(AY)$ equals to $vol_k$ and can be computed by the $k$-th diagonal element in $\{1\}^{r\times n}AY$. 
Then let $trace(\cdot)$ be the trace of input matrix,
\begin{equation}
\notag
\begin{small}
   \begin{aligned}
        NPSI(A,Y) &= NPSI_{\gG P(\gG)}\\
        =&\sum_{k<r}\left(\frac{vol_k-g_k}{2|\gE|}\log_2\frac {vol_k}{2|\gE|}\right)\\
        =&\sum_{k<r}\left(\frac{(Y^TAY)_{kk}}{2 sum(A)} \times \log_2\left(\frac {(\{1\}^{r\times n}AY)_{kk}}{2 sum(A)}\right)\right)\\
        =&\trace\left(\frac{Y^TAY}{2 sum(A)}\otimes \log_2\left(\frac{\{1\}^{r\times n}AY}{2 sum(A)}\right)\right)
    \end{aligned}
\end{small}
\end{equation}
With the definition above, $NPSI(A,Y)$ can be incorporated into GNN.
NPSI is desigend to learn $Y$ on a fixed $\gG$ \cite{li2016structural}.
However, if we fix an $Y\in \{0,1\}^{n\times r}$ which satisfies $(Y^TY)_{kk} > 0$ and $(Y^TY)_{km} = 0$ for  $k\neq m$, we can learn a graph $\gG'$ with corresponding adjacency matrix $A'$ satisfying $Rank(A')\geq r$:

\begin{theorem}[minimize $NPSI$ with learnable $A'$]
    \label{thm:Fix Y to learn graph}
  \begin{equation}
    \label{eq:learn G with Y fixed}
        \begin{aligned}
           \text{Suppose } A' =& \arg\min_{A'}\left( NPSI(A',Y) \right),\\
            s.t.& A'_{ij} \geq 0 \text{ and } A' = A'^T,
        \end{aligned}
    \end{equation}
    $A'$  satisfies:
    $Rank(A')\geq r$
\end{theorem}
Therefore, based on NPSI, if we set $r=c$, we construct an objective function to learn an adjacency $A'$ which satisfies necessary condition Corollary~\ref{cor:Necessary condition of being innocuous graph}.
Besides this, \cite{li2016structural} shows that by minimizing NPSI on fixed $\gG'$,  we can divide the graph into partitions with high inner-connectivity and sparse inter-connectivity.  
Specifically, when input $\gG'$ is fixed, we can obtain the partition of such groups by optimizing:
\begin{equation}
\notag
        \begin{aligned}
            \gC_k=&\{v_i|Y_{ik}\neq 0\}\text{ where, }\\
            Y = & \arg\min_{Y}\left( NPSI(A',Y) \right)\\
            s.t.\text{ }& Y\in \{0,1\}^{n\times r},\text{ } (Y^TY)_{km}\begin{cases}
                     > 0 & \text{ if $k=m$,} \\ 
                     = 0 & \text{ otherwise.}
                \end{cases}
        \end{aligned}
    \end{equation}
With the partition indicator $Y$, we utilize the well known Davies-Bouldin index (DBI) to analyze the similarity of node features inside same group \cite{davies1979cluster}:
\begin{equation}       
        \begin{small}
		\begin{aligned}
			DBI(X,Y)&=\frac{1}{r}\sum_{k<r} DI_k \\
			\text{ where: }  DI_k&=max_{m\neq k}(R_{km})\text{ , }R_{km}=\frac{S_k+S_m}{M_{km}}\\ S_k&=(\frac{1}{|\gC_k|}\sum_{Y_{ik}=1}(|X_i-\overline{X}_k|^2))^{\frac{1}{2}},\\ M_{km}&=(|\overline{X}_k-\overline{X}_m|^2)^{\frac{1}{2}}\text{ , } \overline{X}_k=\frac{\sum_{Y_{ik}=1}X_i}{|\gC_k|}.\\
		\end{aligned}
		\end{small}
	\end{equation}
An adjacency matrix $A'$ satisfies Assumptions~\ref{assump:Group-level feature smoothness} would make $DBI(X,Y)$ small.
Therefore based on NPSI, we construct an objective function to learn an adjacency $A$ which satisfies the necessary conditions (Corollary~\ref{cor:Necessary condition of being innocuous graph} and Assumptions~\ref{assump:Group-level feature smoothness}) simultaneously.
Let $\beta$ be a hyper-parameter. The objective function is:
\begin{equation}       
\label{eq:NPSI loss}
	\begin{aligned}
		\mathcal{L}_N&=NPSI(A',Y) + \beta DBI(X,Y)\\
		s.t.\text{ }& A'_{ij} \geq 0,\text{ } A' = A'^T,\\
		            & Y\in \{0,1\}^{n\times c},\text{ }  Y_{km}\begin{cases}
                                                                 > 0 & \text{ if $k=m$,} \\ 
                                                                 = 0 & \text{ otherwise.}
                                                            \end{cases}
	\end{aligned}
\end{equation}
Then our overall criterion for finding a innocuous graph is formulated into an optimization problem of minimizing $\mathcal{L}_N$ in (\ref{eq:NPSI loss}), where $Y$ and $A'$ are elements to be optimized.

\section{Unsupervised Structural Entropy-based Robust Graph Neural Network}
\label{sec:USER-structure}
In this section, we propose new framework that facilitates GNN models to learn embeddings and innocuous graph simultaneously. 
This framework accomplishes robust learning task by optimizing loss in (\ref{eq:NPSI loss}).
Here we take classical GAE \cite{kipf2016variational} as supported GNN model.
We introduce it from two aspects: structure and optimization.

\textbf{Structure}.
Let $A$ denote the adjacency matrix of original input graph.
To remove the effective of randomness, we construct an innocuous graph and use it as the input of the supported model instead of the original graph. We thus construct a learnable matrix $A'\in\mathbb{R}^{n\times n}$, and use it as the input of supported GNN model:
\begin{equation}       
    \label{eq:USER structure}
	\begin{aligned}
		&H=\GNN\left(A',X,\left\{W^{(1)},W^{(2)}\right\}\right)\\
	\end{aligned}
\end{equation}
$H$ is the learnt node embeddings.
Besides the node embeddings, we add a softmax layer with learnable parameter matrix $W^Y\in \mathbb{R}^{d^(2)\times c}$ to obtain the group indicator matrix $Y$:
\begin{equation}       
    \label{eq:USER obtain Y}
	\begin{aligned}
 		Y=softmax(H W^Y)
	\end{aligned}
\end{equation}

\textbf{Optimization}.
Let $\mathcal{L}_{S}$ be the loss function of supported model, e.g., for GAE:
\begin{equation}       
		\begin{aligned}
			\mathcal{L}_{S}=&||\hat{A}-A||_F^2,
		\end{aligned}
	\end{equation}
where $\hat{A}$ is reconstructed from learnt node embeddings by $\hat{A}=sigmoid(HH^T)$.
Besides $\mathcal{L}_{S}$, $\mathcal{L}_N$ in (\ref{eq:NPSI loss}) is employed to alleviate the interference of randomness.
Thus, let $\alpha$ be hyper-parameter, model is trained by minimizing $\mathcal{L}$:
\begin{equation}       
		\begin{aligned}
			\mathcal{L}=\mathcal{L}_N + \alpha\mathcal{L}_S.
		\end{aligned}
	\end{equation}
Although unsupervised, with structural entropy based $\mathcal{L}_N$, this framework mitigate randomness-interference, making the supported model more capable.
We call it Unsupervised Structural Entropy-based Robust Graph Neural Network (USER).
The detailed structure is shown in Figure \ref{The Structure of USER}.

\begin{table}[htb]
\caption{Dataset statistics.}
\label{Dataset statistics.}
\begin{center}
\scalebox{0.8}{
\begin{tabular}{lrrrr}

\hline Dataset & \# Nodes & \# Edges & \# Features & \# Classes \\
\hline Cora & 2,708 & 5,429 & 1,433 & 7 \\
Citeseer & 3,327 & 4,732 & 3,703 & 6 \\
Wiki & 2,405 & 17,981 & 4,973 & 17 \\
\hline
\end{tabular}
}
\end{center}
\end{table}
\begin{table*}[htb]
\caption{Node clustering performance (NMI±Std) under random-noises}
\label{tab:Node clustering performance (NMI±Std) under random-noises}
\begin{center}
\scalebox{0.57}{
\begin{tabular}{l|c|ccccccccccc}
\hline 
Dataset & Ptb Rate (\%) &deepwalk  &GAE  &VGAE  &ARGA  &AGE  &DGI  &GIC  &GCA  &GAE\_CG  &ARGA\_CG  &USER \\
\hline 
  & 0 & $39.58\pm2.17$ & $44.32\pm3.46$ & $43.42\pm4.78$ & $44.12\pm3.24$ & $56.4\pm3.13$ & $\mathbf{57.32\pm1.02}$ & $52.16\pm0.94$ & $32.76\pm4.35$ & $44.3\pm2.69$ & $45.18\pm4.41$ & $56.24\pm1.58$\\
  & 10 & $35.2\pm2.3$ & $41.28\pm2.11$ & $39.65\pm3.55$ & $40.05\pm2.32$ & $42.23\pm0.55$ & $53.45\pm1.0$ & $50.24\pm0.69$ & $35.1\pm2.33$ & $42.47\pm1.87$ & $43.11\pm2.6$ & $\mathbf{54.38\pm2.23}$\\
  & 20 & $28.61\pm1.25$ & $33.0\pm2.91$ & $35.16\pm2.43$ & $34.6\pm3.11$ & $35.1\pm0.61$ & $50.47\pm0.78$ & $48.54\pm0.69$ & $32.18\pm3.05$ & $38.31\pm3.07$ & $37.96\pm1.89$ & $\mathbf{52.17\pm1.94}$\\
cora & 30 & $26.55\pm2.84$ & $29.07\pm4.38$ & $32.31\pm1.51$ & $29.49\pm4.1$ & $36.05\pm1.04$ & $48.22\pm0.48$ & $44.75\pm0.57$ & $32.93\pm2.23$ & $35.19\pm1.99$ & $36.59\pm1.47$ & $\mathbf{52.39\pm1.0}$\\
  & 40 & $21.21\pm1.57$ & $27.08\pm2.53$ & $27.35\pm2.65$ & $26.86\pm2.55$ & $32.68\pm0.63$ & $44.02\pm1.26$ & $40.97\pm0.96$ & $33.4\pm1.73$ & $33.47\pm2.38$ & $34.46\pm2.23$ & $\mathbf{46.7\pm2.7}$\\
  & 50 & $21.5\pm2.4$ & $25.03\pm3.05$ & $24.99\pm3.27$ & $23.71\pm2.64$ & $36.79\pm2.81$ & $43.22\pm0.69$ & $40.73\pm1.0$ & $31.81\pm1.65$ & $31.93\pm1.86$ & $31.53\pm2.72$ & $\mathbf{49.33\pm1.95}$\\
\hline 
  & 0 & $13.89\pm1.32$ & $21.66\pm3.95$ & $20.84\pm5.63$ & $20.72\pm3.03$ & $35.82\pm0.89$ & $\mathbf{44.02\pm0.57}$ & $43.56\pm0.65$ & $28.1\pm2.89$ & $21.3\pm3.38$ & $19.75\pm3.99$ & $35.52\pm3.51$\\
  & 10 & $11.58\pm1.74$ & $18.3\pm2.69$ & $17.52\pm4.41$ & $17.94\pm2.97$ & $29.47\pm1.85$ & $\mathbf{41.31\pm0.72}$ & $41.29\pm0.76$ & $10.75\pm2.2$ & $20.41\pm1.9$ & $18.26\pm2.54$ & $37.04\pm1.46$\\
  & 20 & $8.77\pm0.6$ & $16.16\pm1.62$ & $15.07\pm2.59$ & $15.6\pm2.78$ & $21.65\pm0.74$ & $36.66\pm0.71$ & $\mathbf{36.72\pm0.91}$ & $7.0\pm0.65$ & $16.84\pm1.77$ & $16.93\pm1.19$ & $34.42\pm3.35$\\
citeseer & 30 & $7.58\pm1.24$ & $12.86\pm1.75$ & $13.59\pm2.22$ & $13.01\pm1.55$ & $18.06\pm1.22$ & $33.39\pm0.78$ & $33.58\pm0.95$ & $6.16\pm1.3$ & $15.17\pm1.63$ & $14.41\pm1.37$ & $\mathbf{34.5\pm3.06}$\\
  & 40 & $6.85\pm1.03$ & $9.81\pm2.52$ & $10.34\pm0.9$ & $10.03\pm1.38$ & $15.57\pm1.11$ & $32.26\pm0.73$ & $31.91\pm0.89$ & $4.22\pm0.88$ & $12.71\pm1.4$ & $12.03\pm2.31$ & $\mathbf{34.58\pm2.49}$\\
  & 50 & $5.49\pm0.58$ & $10.41\pm1.73$ & $10.33\pm1.07$ & $9.63\pm1.29$ & $14.13\pm0.91$ & $29.58\pm0.94$ & $30.96\pm0.77$ & $2.98\pm0.35$ & $12.59\pm2.83$ & $13.66\pm1.71$ & $\mathbf{34.5\pm2.1}$\\
\hline 
  & 0 & $35.85\pm1.25$ & $23.52\pm8.9$ & $24.49\pm5.63$ & $22.8\pm7.73$ & $\mathbf{51.2\pm1.94}$ & $43.14\pm1.12$ & $27.45\pm1.69$ & $37.54\pm1.05$ & $22.72\pm11.1$ & $23.04\pm8.6$ & $48.99\pm1.16$\\
  & 10 & $33.15\pm1.19$ & $22.57\pm8.43$ & $16.59\pm7.4$ & $18.01\pm8.58$ & $48.85\pm1.01$ & $40.13\pm0.71$ & $24.65\pm1.43$ & $30.77\pm1.62$ & $19.69\pm8.96$ & $19.63\pm9.06$ & $\mathbf{48.97\pm1.16}$\\
  & 20 & $30.74\pm1.31$ & $13.66\pm7.96$ & $14.22\pm6.89$ & $14.05\pm7.05$ & $46.92\pm0.57$ & $36.15\pm1.36$ & $22.9\pm1.66$ & $30.74\pm1.09$ & $11.62\pm5.32$ & $16.54\pm7.83$ & $\mathbf{48.71\pm1.63}$\\
wiki & 30 & $27.79\pm0.86$ & $14.7\pm4.2$ & $15.97\pm6.93$ & $15.1\pm5.56$ & $47.43\pm0.65$ & $34.84\pm0.68$ & $38.29\pm0.68$ & $31.42\pm1.64$ & $16.5\pm6.29$ & $10.66\pm5.27$ & $\mathbf{48.55\pm1.44}$\\
  & 40 & $26.56\pm1.52$ & $8.26\pm7.51$ & $9.8\pm6.17$ & $15.0\pm4.93$ & $46.7\pm0.59$ & $31.28\pm1.25$ & $36.33\pm0.54$ & $32.38\pm1.39$ & $9.99\pm10.57$ & $11.41\pm6.2$ & $\mathbf{48.54\pm2.02}$\\
  & 50 & $25.52\pm1.27$ & $9.52\pm6.89$ & $7.4\pm6.42$ & $12.01\pm5.11$ & $46.83\pm0.58$ & $29.29\pm1.15$ & $33.26\pm0.89$ & $34.24\pm1.75$ & $9.26\pm4.87$ & $6.6\pm3.32$ & $\mathbf{48.68\pm1.78}$\\
\hline
\end{tabular}
}
\end{center}
\end{table*}

\begin{table*}[htb]
\caption{Node clustering performance (NMI±Std) under meta-attack}
\label{tab:Node clustering performance (NMI±Std) under meta-attack}
\begin{center}
\scalebox{0.57}{
\begin{tabular}{l|c|ccccccccccc}
\hline 
Dataset & Ptb Rate (\%) &deepwalk  &GAE  &VGAE  &ARGA  &AGE  &DGI  &GIC  &GCA  &GAE\_CG  &ARGA\_CG  &USER \\
\hline 
  & 5 & $41.73\pm2.16$ & $43.37\pm3.34$ & $43.06\pm2.66$ & $43.33\pm3.28$ & $48.6\pm1.73$ & $50.33\pm2.3$ & $46.89\pm2.05$ & $38.12\pm3.46$ & $43.64\pm3.44$ & $43.0\pm3.15$ & $\mathbf{50.64\pm2.77}$\\
  & 10 & $37.68\pm2.86$ & $34.1\pm3.44$ & $33.6\pm3.66$ & $34.5\pm3.71$ & $39.35\pm3.14$ & $37.73\pm3.63$ & $36.58\pm3.11$ & $34.07\pm2.77$ & $35.47\pm2.79$ & $35.94\pm3.51$ & $\mathbf{41.71\pm3.32}$\\
cora & 15 & $21.99\pm4.38$ & $19.96\pm4.11$ & $19.56\pm4.18$ & $20.04\pm3.82$ & $25.39\pm3.88$ & $23.13\pm3.39$ & $23.19\pm3.29$ & $21.54\pm4.61$ & $22.59\pm3.69$ & $22.92\pm3.38$ & $\mathbf{29.27\pm3.68}$\\
  & 20 & $7.31\pm2.85$ & $7.26\pm2.86$ & $7.22\pm2.91$ & $7.88\pm2.79$ & $9.65\pm3.35$ & $10.17\pm2.67$ & $10.96\pm3.11$ & $9.97\pm2.01$ & $10.34\pm3.2$ & $10.31\pm2.96$ & $\mathbf{18.82\pm2.9}$\\
\hline 
  & 5 & $16.97\pm2.14$ & $22.5\pm3.43$ & $22.73\pm2.68$ & $20.85\pm2.69$ & $34.06\pm1.99$ & $\mathbf{40.22\pm1.89}$ & $39.91\pm1.95$ & $20.78\pm5.93$ & $22.67\pm2.54$ & $21.69\pm3.07$ & $35.72\pm2.03$\\
  & 10 & $23.52\pm1.69$ & $22.25\pm2.6$ & $22.59\pm2.62$ & $22.02\pm2.22$ & $25.13\pm2.7$ & $29.71\pm3.05$ & $29.45\pm3.0$ & $18.92\pm1.91$ & $22.6\pm1.94$ & $22.06\pm1.87$ & $\mathbf{31.86\pm2.84}$\\
citeseer & 15 & $17.33\pm2.73$ & $13.73\pm2.81$ & $13.6\pm2.95$ & $13.94\pm2.65$ & $15.71\pm2.01$ & $17.68\pm2.77$ & $17.81\pm2.68$ & $13.61\pm1.99$ & $15.6\pm2.35$ & $15.61\pm2.15$ & $\mathbf{27.77\pm3.31}$\\
  & 20 & $8.3\pm2.45$ & $5.64\pm2.01$ & $5.71\pm1.82$ & $5.63\pm1.74$ & $9.11\pm0.85$ & $9.11\pm1.78$ & $9.08\pm2.17$ & $7.08\pm2.03$ & $7.68\pm2.06$ & $7.61\pm2.12$ & $\mathbf{26.42\pm2.67}$\\
\hline 
  & 5 & $34.06\pm1.74$ & $19.59\pm7.49$ & $19.22\pm7.6$ & $20.8\pm5.9$ & $41.76\pm1.31$ & $32.94\pm2.61$ & $35.03\pm3.18$ & $27.24\pm1.4$ & $18.24\pm7.97$ & $16.27\pm5.05$ & $\mathbf{48.44\pm1.71}$\\
  & 10 & $22.96\pm2.74$ & $13.09\pm6.62$ & $11.14\pm5.52$ & $12.48\pm4.46$ & $38.72\pm0.26$ & $22.59\pm3.1$ & $23.64\pm2.65$ & $25.86\pm1.81$ & $13.34\pm6.63$ & $10.98\pm4.41$ & $\mathbf{47.71\pm1.7}$\\
wiki & 15 & $14.35\pm1.8$ & $4.59\pm5.02$ & $4.99\pm4.31$ & $6.82\pm3.15$ & $40.9\pm0.89$ & $12.27\pm2.43$ & $15.19\pm0.93$ & $20.14\pm5.08$ & $4.62\pm5.21$ & $7.04\pm3.79$ & $\mathbf{47.54\pm1.53}$\\
  & 20 & $9.3\pm1.2$ & $2.22\pm3.4$ & $1.52\pm3.09$ & $3.61\pm1.85$ & $42.71\pm0.98$ & $8.85\pm0.77$ & $9.24\pm2.33$ & $15.39\pm2.7$ & $3.1\pm3.9$ & $2.49\pm0.21$ & $\mathbf{47.48\pm1.54}$\\
\hline
\end{tabular}
}
\end{center}
\end{table*}

\begin{table}[htb]
\caption{Link prediction (AUC±Std) under random-noises}
\label{Link prediction performance (AUC±Std) under edge-flip noises}
\begin{center}
\scalebox{0.55}{
\begin{tabular}{l|c|ccccc}
\hline 
Dataset & Ptb Rate (\%) &GAE  &ARGA  &GAE\_CG  &ARGA\_CG  &USER \\
\hline 
  & 0 & $94.09\pm0.26$ & $94.87\pm0.48$ & $94.08\pm0.63$ & $94.0\pm0.88$ & {$\mathbf{95.38\pm0.12}$}\\
  & 10 & $94.09\pm0.78$ & $94.25\pm0.21$ & $93.96\pm0.41$ & $94.04\pm1.03$ & $\mathbf{95.59\pm0.16}$\\
  & 20 & $94.12\pm0.09$ & $93.69\pm1.16$ & $94.01\pm0.44$ & $94.05\pm0.34$ & $\mathbf{94.99\pm0.65}$\\
citeseer & 30 & $91.72\pm0.2$ & $92.31\pm0.82$ & $93.15\pm0.1$ & $93.27\pm0.17$ & $\mathbf{95.15\pm0.17}$\\
  & 40 & $90.29\pm2.99$ & $90.81\pm0.53$ & $93.07\pm0.79$ & $92.89\pm0.46$ & $\mathbf{94.41\pm0.21}$\\
  & 50 & $90.05\pm1.95$ & $90.91\pm0.51$ & $91.85\pm0.87$ & $91.6\pm0.12$ & $\mathbf{94.54\pm0.24}$\\
\hline 
  & 0 & $86.75\pm1.05$ & $82.14\pm15.09$ & $83.25\pm7.29$ & $79.81\pm6.89$ & $\mathbf{88.72\pm0.14}$\\
  & 10 & $80.12\pm16.61$ & $83.66\pm6.24$ & $68.04\pm12.66$ & $77.24\pm4.86$ & ${88.07\pm0.32}$\\
  & 20 & $79.5\pm4.86$ & $80.86\pm10.5$ & $70.62\pm6.33$ & $74.57\pm3.44$ & $\mathbf{87.82\pm0.27}$\\
wiki & 30 & $73.02\pm10.44$ & $80.06\pm3.96$ & $66.27\pm9.57$ & $68.97\pm6.42$ & $\mathbf{87.41\pm0.51}$\\
  & 40 & $78.37\pm4.91$ & $79.44\pm4.46$ & $61.58\pm6.22$ & $64.06\pm9.79$ & $\mathbf{87.45\pm0.22}$\\
  & 50 & $67.78\pm4.55$ & $76.79\pm2.16$ & $64.48\pm9.48$ & $71.51\pm8.56$ & $\mathbf{86.87\pm0.4}$\\
\hline
\end{tabular}
}
\end{center}
\end{table}
\begin{table}[htb]
	\caption{Link prediction (AP±Std) under random-noises}
	\label{Link prediction performance (AP±Std) under edge-flip noises}
	\begin{center}
		\scalebox{0.55}{
			\begin{tabular}{l|c|ccccc}
				\hline 
				Dataset & Ptb Rate (\%) &GAE  &ARGA  &GAE\_CG  &ARGA\_CG  &USER \\
				\hline 
				& 0 & $94.22\pm0.32$ & $94.89\pm0.5$ & $94.5\pm0.29$ & $94.14\pm0.74$ & {{$\mathbf{95.84\pm0.08}$}}\\
				& 10 & $94.08\pm0.94$ & $94.47\pm0.44$ & $94.21\pm0.41$ & $94.38\pm1.19$ & {{$\mathbf{95.98\pm0.08}$}}\\
				& 20 & $94.57\pm0.01$ & $94.05\pm1.14$ & $94.5\pm0.24$ & $94.42\pm0.51$ & {{$\mathbf{95.46\pm0.67}$}}\\
				citeseer & 30 & $92.13\pm0.21$ & $92.77\pm0.93$ & $93.83\pm0.32$ & $93.94\pm0.17$ & {{$\mathbf{95.66\pm0.17}$}}\\
				& 40 & $90.95\pm2.81$ & $91.38\pm0.33$ & $93.62\pm0.61$ & $93.58\pm0.25$ & {{$\mathbf{94.92\pm0.16}$}}\\
				& 50 & $90.81\pm2.12$ & $91.66\pm0.63$ & $92.73\pm0.71$ & $92.44\pm0.28$ & {{$\mathbf{95.05\pm0.19}$}}\\
				\hline 
				& 0 & $88.18\pm1.49$ & $83.38\pm17.68$ & $85.65\pm7.11$ & $83.01\pm6.33$ & {{$\mathbf{89.9\pm0.1}$}}\\
				& 10 & $81.95\pm17.34$ & $86.02\pm4.96$ & $69.8\pm14.58$ & $80.78\pm4.35$ & {{$\mathbf{89.48\pm0.22}$}}\\
				& 20 & $82.62\pm3.11$ & $82.44\pm12.9$ & $73.09\pm6.73$ & $77.22\pm3.33$ & {{$\mathbf{89.07\pm0.27}$}}\\
				wiki & 30 & $76.89\pm12.98$ & $82.98\pm2.6$ & $69.06\pm11.63$ & $72.13\pm6.08$ & {{$\mathbf{88.9\pm0.27}$}}\\
				& 40 & $81.53\pm3.1$ & $82.22\pm2.94$ & $63.15\pm7.47$ & $65.91\pm12.33$ & {{$\mathbf{88.61\pm0.08}$}}\\
				& 50 & $70.36\pm2.81$ & $80.29\pm1.83$ & $67.09\pm10.51$ & $73.6\pm11.85$ & {{$\mathbf{88.28\pm0.38}$}}\\
				\hline
			\end{tabular}
		}
	\end{center}
\end{table}
\begin{figure}[htb]
\centering
\subfigure[Original]{
\begin{minipage}[t]{0.33\linewidth}
\label{The original input graph}
\centering
\includegraphics[width=1.1in]{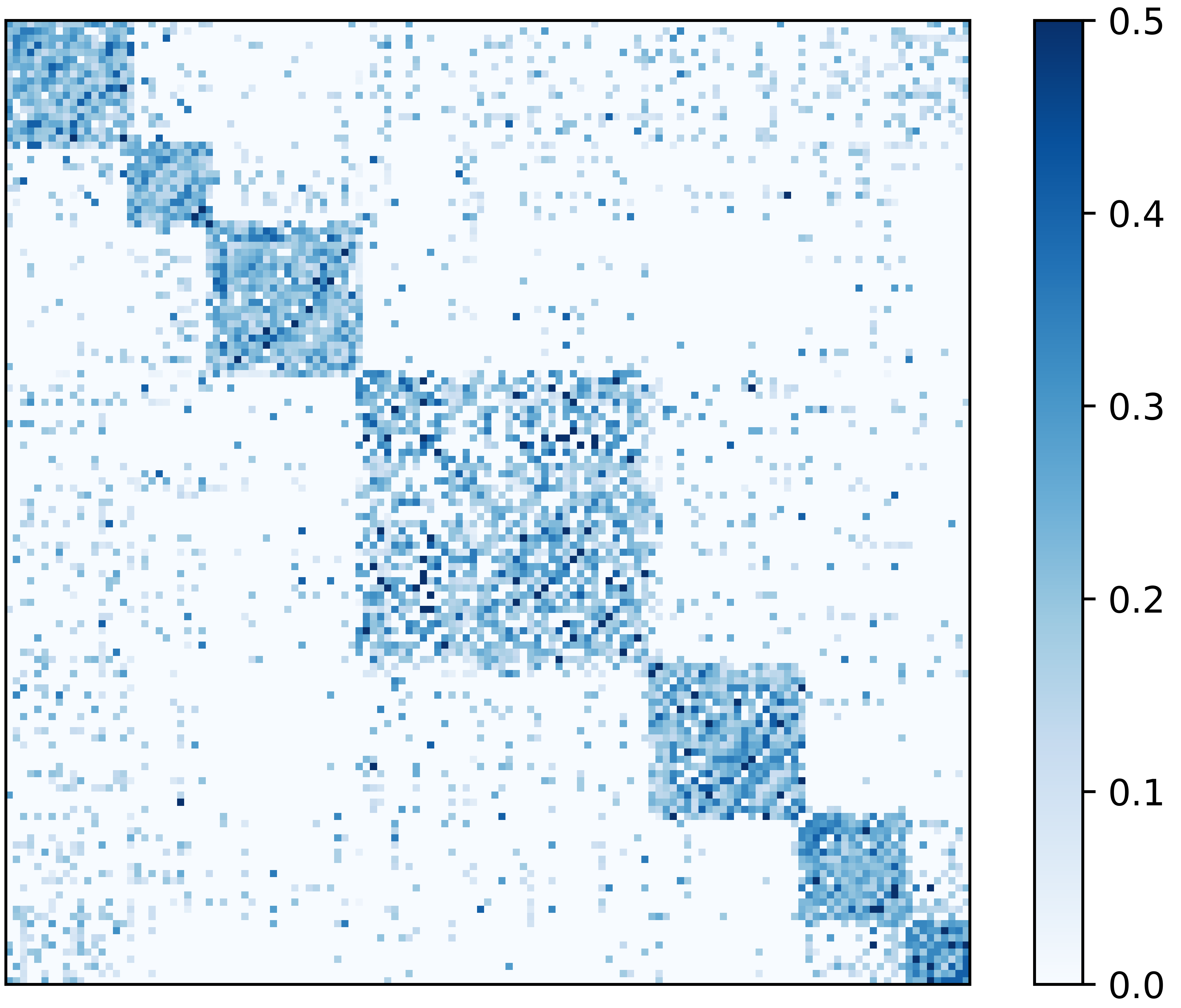}
\end{minipage}%
}%
\subfigure[Poisoned]{
\begin{minipage}[t]{0.33\linewidth}
\label{The input graph poisoned by 50 flip noise}
\centering
\includegraphics[width=1.11in]{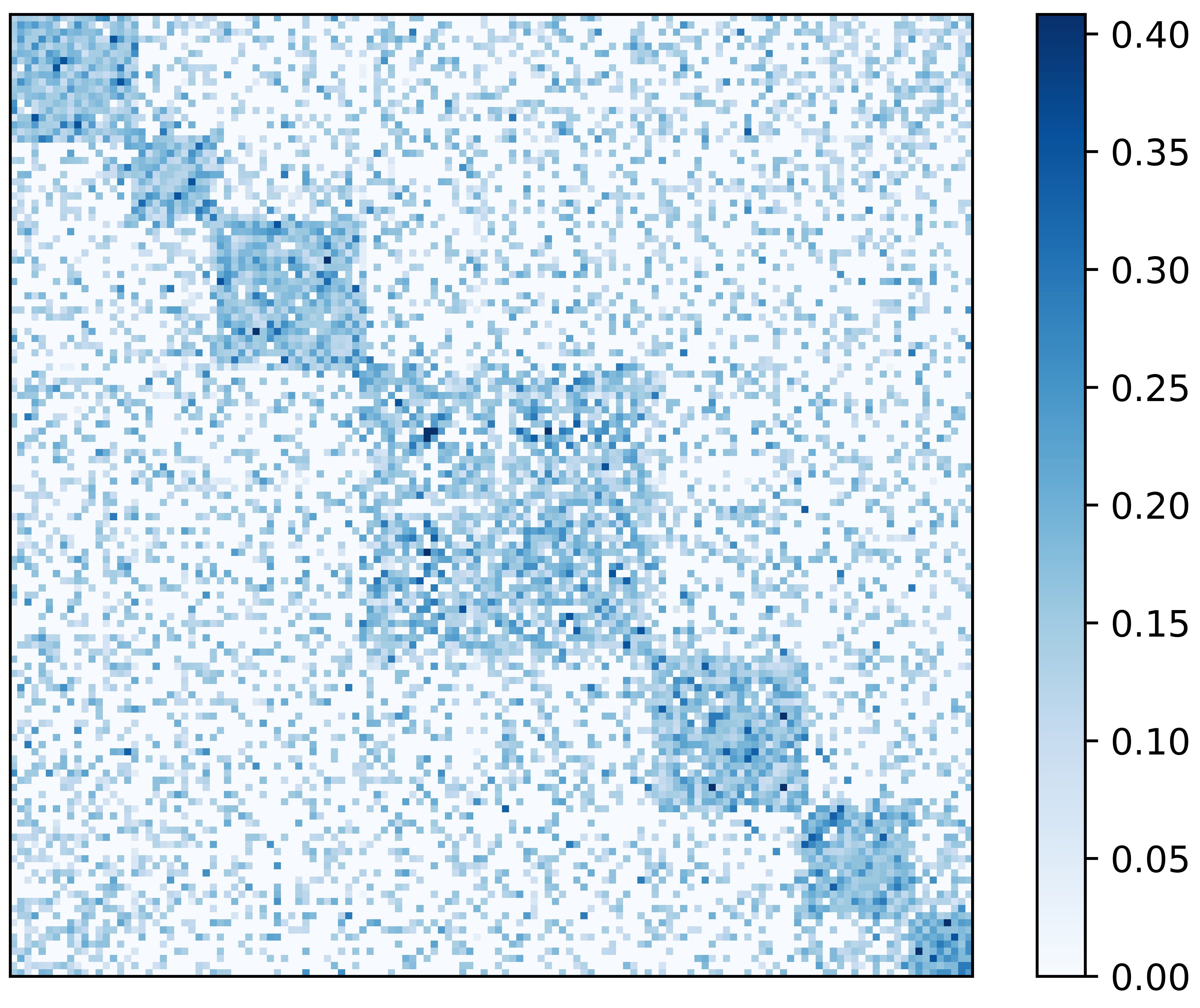}
\end{minipage}%
}%
\subfigure[USER]{
\begin{minipage}[t]{0.33\linewidth}
\label{The output graph learned by USER}
\centering
\includegraphics[width=1.11in]{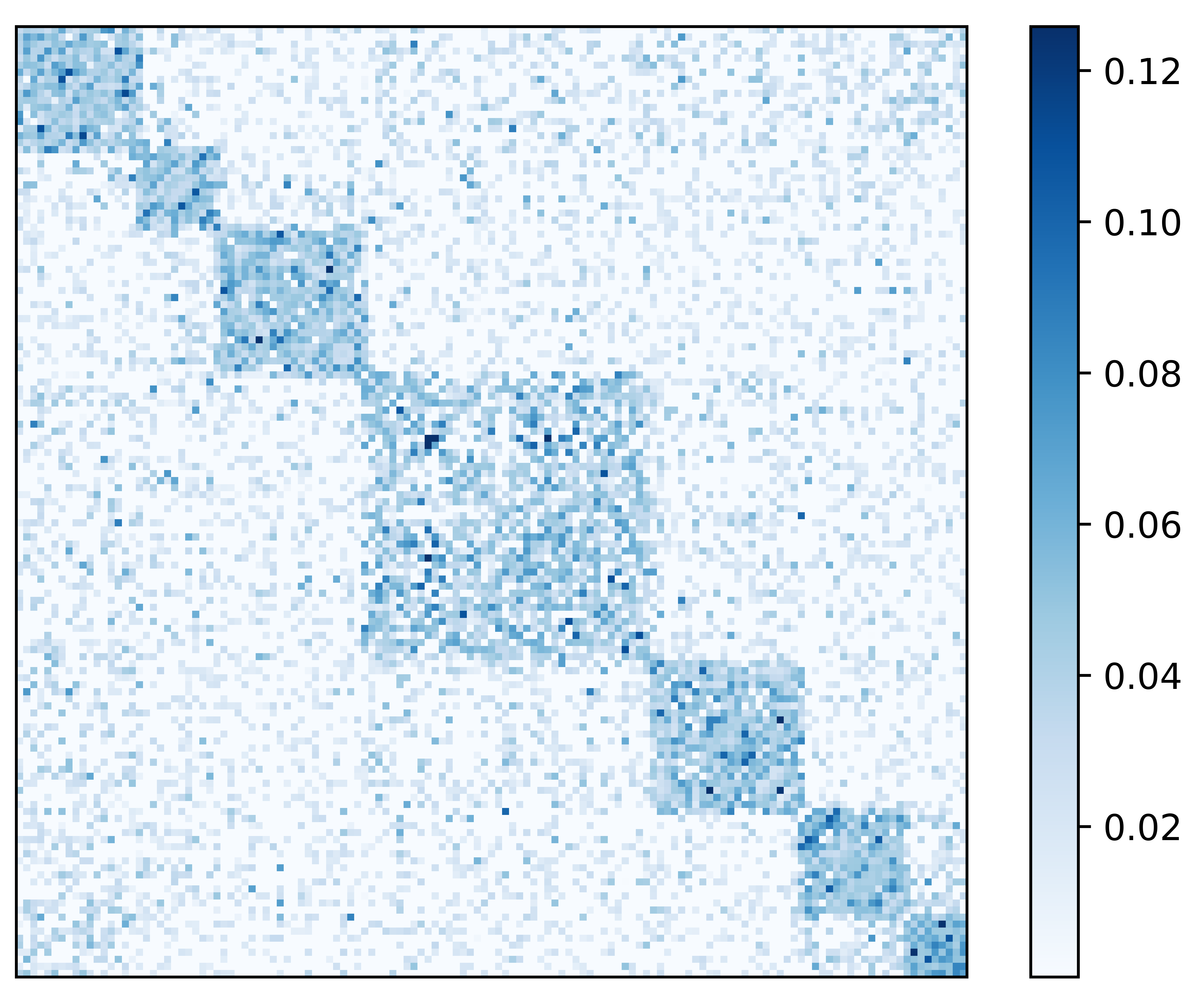}
\end{minipage}
}%
\centering
\caption{Case study: the graph heat maps of Cora}
\end{figure}


\begin{table}[htb]
\caption{NMI of USER's variants with $10\%$ random-noise}
\label{Clustering performance of USER's variants}
\begin{center}
\scalebox{0.8}{
\begin{tabular}{l|cccc}
\hline 
Dataset & USER &w.o. NPSI  &w.o. DBI &Fix $A'$\\
\hline 
 cora & 54.38 &14.82  &52.54 &40.11\\
 citeseer & 37.04 &28.95  &12.82 &30.94\\
 wiki & 48.97 &48.44  &37.28 &39.77\\
\hline
\end{tabular}
}
\end{center}
\end{table}

\begin{figure}[htb]
\centering
\subfigure[$\alpha$ on wiki]{
\begin{minipage}[t]{0.5\linewidth}
\label{Wiki alpha}
\centering
\includegraphics[width=1.8in]{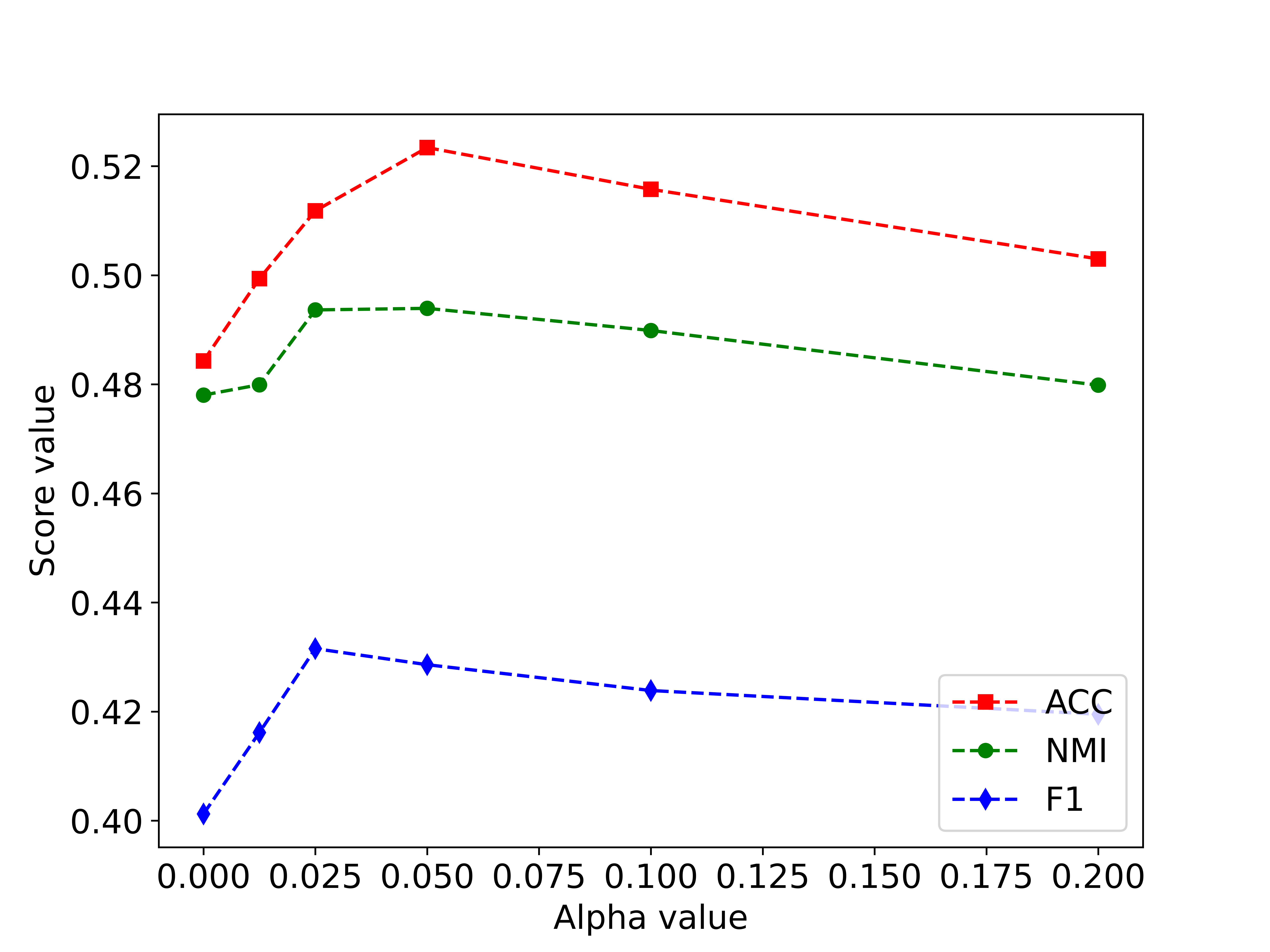}
\end{minipage}
}%
\subfigure[$\beta$ on wiki]{
\begin{minipage}[t]{0.5\linewidth}
\label{Wiki beta}
\centering
\includegraphics[width=1.8in]{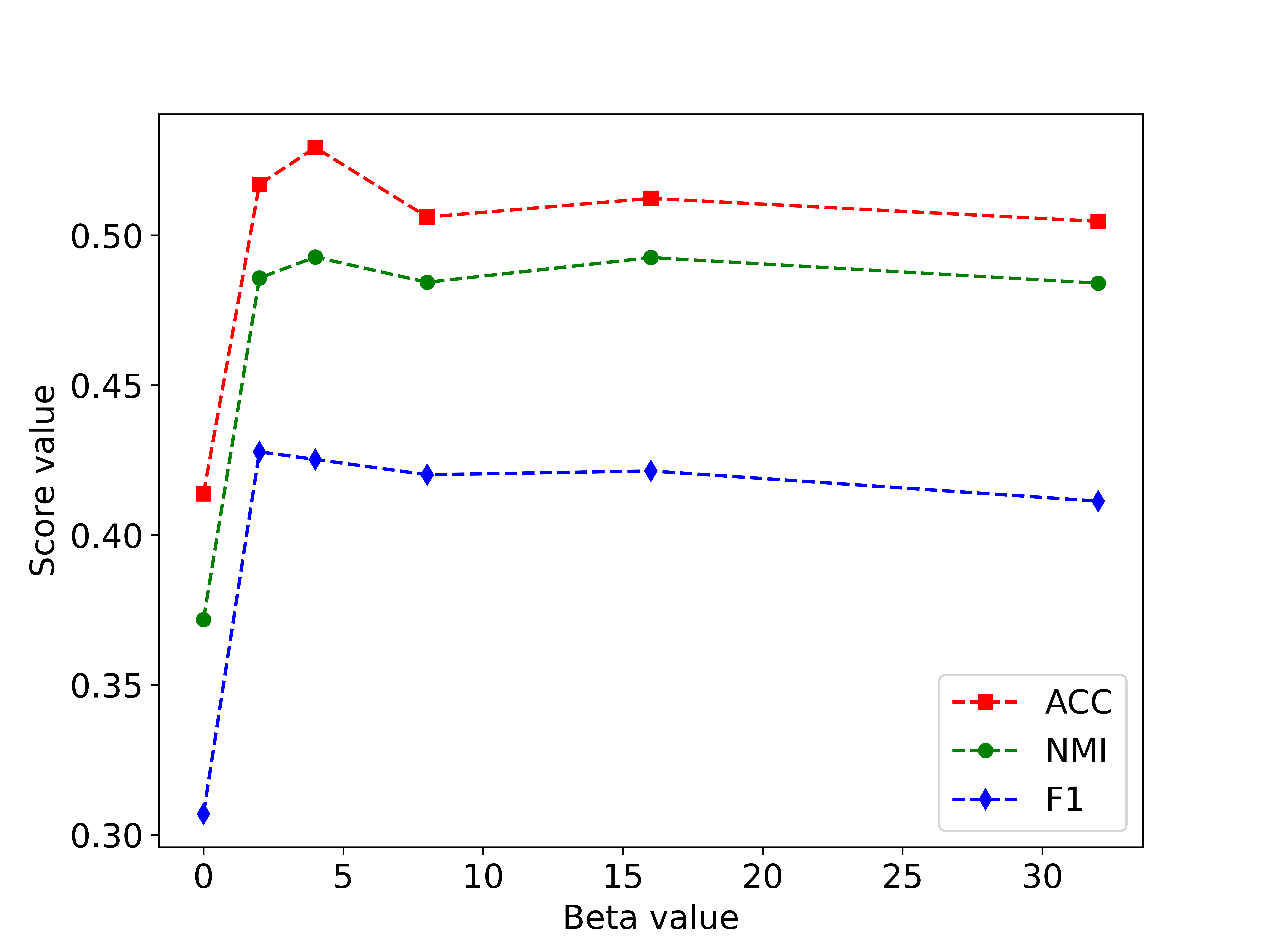}
\end{minipage}
}%
\centering
\caption{Parameter analysis on Citeseer and Wiki}
\label{alpha and beta}
\end{figure}


\section{Experiments}

In this section, we provide the experiments comparing the performance of USER supported GAE (denoted by USER) with other state-of-the-art baseline methods, 
case study, ablation study and parameter analysis.

\subsection{Experimental Settings}

\textbf{Datasets}.
We evaluate all models on three widely-used benchmark datasets: \textit{Cora}, \textit{Citeseer}, \textit{Wiki} \cite{kipf2017semi,  yang2015network,  jin2020graph}. \textit{Cora} and \textit{Citeseer} are citation networks where nodes represent publications and edges stand for citation links. Their node features are the bag-of-words vectors; \textit{Wiki} is a webpage network in which nodes are web pages, and edges represent hyperlinks. The node features in it are  tf-idf weighted
word vectors. The statistics of these datasets are in Table \ref{Dataset statistics.}.

\noindent \textbf{Noises}.
Besides the original graph in datasets, we inject noises into the graph to promote the graph randomness. In particular, we develop two types of noises: {\em random noise} and {\em meta-attack noise} \cite{zugner2019adversarial}. 
Random noise ``randomly flips'' the state of the  chosen pair of nodes (i.e., if there is an edge between them, we remove it; otherwise we add an edge between them). The number of changed edges is the ratio of the total number of edges in original graph. In most cases, random noises are not very effective, so we create several poisoned graphs with noise ratio from $0\%$ to $50\%$ with a step of $10\%$.
Meta-attack noise can promote the graph randomness significantly  \cite{zugner2019adversarial,jin2020adversarial}. 
Even for supervised models, meta-attack is hardly applied with  a perturbation rate higher than $20\%$ \cite{jin2020adversarial}.
Thus, we create several poisoned graphs with meta-attack noise ratio from $0\%$ to $20\%$ with a step of $5\%$.

\noindent \textbf{Baselines}.
For \textbf{USER}, we use classical GAE \cite{kipf2016variational} as its supported model. 
To evaluate the effectiveness, we compare it with $10$ baselines retaining the default parameter settings in their original papers.
\textbf{DeepWalk} \cite{perozzi2014deepwalk} utilizes random walks to learn embeddings.
\textbf{GAE} and \textbf{VGAE}  \cite{kipf2016variational} firstly leverage GCN \cite{kipf2017semi}for GRL. 
\textbf{ARGA} \cite{PanHLJYZ18} is an adversarial GNN model.
\textbf{AGE} \cite{cui2020adaptive} applies Laplacian smoothing to GNN. 
\textbf{DGI} \cite{velickovic2019deep} trains GNN with MI. 
\textbf{GIC} \cite{mavromatis2020graph} captures cluster-level information.
\textbf{GCA} \cite{zhu2021graph} is a Graph Contrastive learning GNN.
\textbf{GAE\_CG} and \textbf{ARGA\_CG} are Cross-Graph \cite{wang2020cross} models. {GAE\_CG} is the GAE version while ARGA\_CG maintains ARGA encoders.
Please note that GCA and Cross-Graph are also unsupervised robust models.
However we firstly introduce innocuous graph, which make USER more effective.

\noindent \textbf{Parameter Settings} 
We train USER for $400$ epochs using Adam optimizer with a learning rate $\eta$.
The two hyper-parameters \textbf{$\alpha$} and \textbf{$\beta$}, are selected through a grid search regarding performance, a detailed analysis could be found in Subsection ~\ref{subsec:Parameter Analysis}.
The dimension $d^{(1)}$, learning rate $\eta$, \textbf{$\alpha$} and \textbf{$\beta$} are selected accordingly based on the parameter analysis. 

\noindent \textbf{Evaluation Metrics}
For node clustering, we employ popular normalized mutual information (NMI) and clustering accuracy (ACC) \cite{aggarwal2014data}.
For link prediction, we report area under the ROC (AUC) \cite{bradley1997use}, and average precision (AP) \cite{su2015relationship}.

\subsection{Performance}
\label{sec:Performance}
\textbf{Clustering}.
We compare the performance of all models in Table \ref{tab:Node clustering performance (NMI±Std) under random-noises} and Table \ref{tab:Node clustering performance (NMI±Std) under meta-attack} 
All the experiments are conducted $10$ times and the average NMI with standard deviation is reported.
For each dataset, the best performance is in bold. 
From Table~\ref{tab:Node clustering performance (NMI±Std) under random-noises} and Table \ref{tab:Node clustering performance (NMI±Std) under meta-attack}, we observe that: 
\textbf{(1) Original graph.} When the input graph is the original graph, the USER's improvement from GAE  is significant. Different from classical GAE, the performances of USER are always close to the best.
\textbf{(2) Random-noises.} When graph randomness is promoted by random noises, USER outperforms others (including GCA and Cross-Graph). Even under large noise rate e.g., $50\%$, performance of USER only drop $12\%$, $3\%$, $0.6\%$ on Cora, Citeseer and Wiki, compared with original graphs. \textbf{(3) Meta-attack.} Meta-attack seems to be more powerful, making effect of most models drop rapidly. However, USER is still more effective than others.

\noindent \textbf{Link prediction}.
To compare the performances on link prediction tasks. We follow the settings in \cite{kipf2016variational}: take out $5\%$ edges from  Citeseer and Wiki datasets to form the validation set and $10\%$ edges for test set, respectively. 
Then we impose random-noises on the rest of the network.
Classical GRL models such as GAE, ARGA and the corresponding Cross-Graph supported version are used as baselines.
All the experiments are run $10$ times and we report the AUC and AP with standard deviation in Table \ref{Link prediction performance (AUC±Std) under edge-flip noises} and Table \ref{Link prediction performance (AP±Std) under edge-flip noises}. The best performance is in bold. 
From the results, we observe that for link prediction, USER also outperforms other models.
Classical models are rather unstable towards promoted randomness. 
Even robust model Cross-graph's performance drop drastically under large ratio noises (e.g. the ARGA\_CG dropped $3.617\%$ and $10.400\%$ on citeseer and wiki when noise rate is $50\%$).
USER demonstrates stability w.r.t. different noise levels (only $0.881\%$ and $2.085\%$ drop with $50\%$ noise).
It verifies that USER can accomplish different tasks facing graph randomness.

\subsection{Case Study}
\label{sec:Case Study}
To show the graph learned by USER.
We illustrate that the normalized adjacency matrix of Cora dataset without noise and rearranged vertices in Figure \ref{The original input graph}. 
It is clearly observable that most edges are in one of seven groups with few edges between them.
On the other hand, the adjacency matrix with $50\%$-ratio random-noises of Cora (as shown in Figure \ref{The input graph poisoned by 50 flip noise}) have more inter-group edges and the boundaries of classes get visibly blurred.
The learned graph structure by USER is shown in Figure \ref{The output graph learned by USER}. From Figure \ref{The output graph learned by USER}, we observe that the group-boundaries are much clearer.
This demonstrates that USER can capture ideal innocuous graph. 

\subsection{Ablation Study}
\label{subsec:Ablation study}
To understand the importance of different components of our model in denoising, we conduct ablation studies on Cora, Citeseer and Wiki datasets with $10\%$ random-noise.
\textbf{NPSI:} From Table \ref{Clustering performance of USER's variants}, USER without NPSI component loses its effectiveness on all three datasets.
\textbf{DBI:} The performance of USER after removing DBI drops slightly on Cora but it is significantly affected  on Wiki and Citeseer. This implies for these two datasets, feature information is more important.
\textbf{Learnable $A'$:} If we fix $A'$ the same as the original input, the model tends to be disturbed by the graph randomness. 
The experimental result on all datasets show the effect. 
By incorporating all these components, USER can explore for innocuous graph and thus consistently outperforms  baselines.

\subsection{Parameter Analysis}
\label{subsec:Parameter Analysis}
We illustrate the mechanism of USER and explore the sensitivity of the two hyper parameters.
\textbf{$\alpha$} controls the influence of the objective function from supported model and 
\textbf{$\beta$} is used to adjust the influence of Assumption~\ref{assump:Group-level feature smoothness}. 
We vary $\alpha$ from $0.003125$ to $0.1$ and $0.00625$ to $0.2$, $\beta$ from $0.025$ to $0.8$ and $1.0$ to $32.0$ in a $\log$ scale of base $2$ respectively.
We report the experiment results on Wiki with $10\%$ random-noise in Figure \ref{alpha and beta} as similar observations are made in other settings 
As we can see, USER's performance can be boosted when choosing appropriate values for all the hyper-parameters, but performance under values too large or too small drops slightly. This is consistent with our analysis.

\section{Conclusion}
We aim to alleviate the interference of graph randomness and learn appropriate node representations without label information. 
We propose USER, a novel unsupervised robust framework. 
Along designing it, we discovered the fact that there are multiple innocuous graphs with which GNN can learn the appropriate embeddings and
introduced rank of adjacency plays a crucial role in discovering such graphs.
We also introduce structural entropy as a tool to construct objective function to capture innocuous graph.
In the future, we'll explore more about intrinsic connectivities of graph data.

\section*{Acknowledgements}
This research was supported by NSFC (Grant No. 61932002) and
Marsden Fund (21-UOA-219).
The first author and third author are supported by a PhD scholarship from China Scholarship Council.
\bibliography{USER}

\end{document}